\documentclass{article}

\usepackage[nonatbib,preprint]{nips_2018}

\usepackage[utf8]{inputenc} 
\usepackage[T1]{fontenc}    
\usepackage{hyperref}       
\usepackage{url}            
\usepackage{booktabs}       
\usepackage{amsfonts}       
\usepackage{nicefrac}       
\usepackage{microtype}      

\usepackage{xcolor}
\usepackage{graphicx} 
\usepackage{caption}
\usepackage{subcaption}

\title{Importance Weighted Evolution Strategies}

\author{
  Víctor Campos\\
  Barcelona Supercomputing Center\\
  \texttt{victor.campos@bsc.es} \\
  \And
  Xavier Giro-i-Nieto\\
  Universitat Politècnica de Catalunya\\
  \texttt{xavier.giro@upc.edu} \\
  \And
  Jordi Torres\\
  Barcelona Supercomputing Center\\
  \texttt{jordi.torres@bsc.es} \\
}

\begin{document}

\maketitle

\begin{abstract}

Evolution Strategies (ES) emerged as a scalable alternative to popular Reinforcement Learning (RL) techniques, providing an almost perfect speedup when distributed across hundreds of CPU cores thanks to a reduced communication overhead. Despite providing large improvements in wall-clock time, ES is data inefficient when compared to competing RL methods. One of the main causes of such inefficiency is the collection of large batches of experience, which are discarded after each policy update. In this work, we study how to perform more than one update per batch of experience by means of Importance Sampling while preserving the scalability of the original method. The proposed method, Importance Weighted Evolution Strategies (IW-ES), shows promising results and is a first step towards designing efficient ES algorithms.

\end{abstract}
\section{Introduction}

The pace of advances in machine learning is frequently upper bounded by the time taken to train models. Even though hardware manufacturers continuously provide improvements in computational power~\cite{volta_whitepaper}, the community has turned to distributed solutions to further reduce training times~\cite{dean2012_distbelief} and training larger models~\cite{shazeer2017_outrageously}. However, accelerating an algorithm by distributing it across several computing devices is not always a trivial task. The communication overhead precludes the distribution of some methods beyond a reduced number of machines~\cite{chen2016_revisiting, campos2017_distributed}, and sometimes parallel training can even hinder the final performance of the model when done naively~\cite{goyal2017_accurate}. This motivates research efforts towards developing algorithms that are well suited for parallel training, from both learning and computational standpoints.

Evolution Strategies (ES)~\cite{salimans2017_es} were proposed as a scalable alternative to popular Reinforcement Learning techniques. Thanks to a reduced communication overhead, ES can be scaled to over a thousand CPU cores with almost linear speedup, providing massive improvements in wall-clock time when training agents in well-known RL benchmarks. 
However, this speedup comes at the cost of a reduced data efficiency, i.e.~ES need more interactions with the environment to achieve the same score as competing methods. Even though this trade-off might not be problematic for simulated tasks, where one can turn compute into data, data efficiency is crucial for the deployment of RL agents in real world scenario, e.g.~robot manipulation tasks~\cite{riedmiller2018_sac}. Research has been conducted to improve the data efficiency of other RL methods~\cite{schaul2015_prioritized, gu2016_qprop}, and we believe that ES would benefit from similar efforts as well.

We aim at improving the data efficiency of ES, while maintaining the scalability of the original method. Our contributions can be summarized as follows: (1)~we propose Importance Weighted Evolution Strategies (IW-ES), an extension of ES that can perform more than one update per batch of experience, (2)~analyze the scalability of IW-ES from the computational standpoint, and (3)~report preliminary results for IW-ES under different configurations that provide insight on the potential of the method and possible improvements to overcome its current limitations.

\section{Background: Evolution Strategies}

The term \textit{Evolution Strategies} (ES)~\cite{rechenberg1973_evolutionsstrategie} makes reference to a class of black box optimization algorithms which implement heuristics inspired by natural evolution. However, throughout this manuscript, we will use the term to refer to the particular algorithm proposed by Salimans et al.~\cite{salimans2017_es}. This method, which belongs to the class of Natural Evolution Strategies~\cite{wierstra2008_nes, wierstra2014_nes}, was shown to be competitive for solving RL problems while exhibiting some attractive features. These features include invariance to action frequency and reward distribution, the possibility to optimize non-differentiable policies, and being highly parallelizable thanks to an efficient communication strategy.

Let $F$ denote the objective function acting on parameters $\mathbf{\theta}$, defined in RL problems as the stochastic score experienced by an agent after a complete trajectory. 
ES seeks to maximize $\mathbb{E}_{\mathbf{\theta} \sim p_\psi}F(\mathbf{\theta})$, the average objective over a population of solutions $p_\psi$, using the score function estimator for the gradient.
Salimans et al.~\cite{salimans2017_es} instantiate the population as a multivariate Gaussian with diagonal covariance matrix centered at $\mathbf{\theta}$, thus obtaining the following estimator:
\begin{equation}
    \nabla_\mathbf{\theta} \mathop{\mathbb{E}_{\mathbf{\epsilon} \sim N(0, \sigma^2 \mathbf{I})}} F(\mathbf{\theta} + \mathbf{\epsilon}) = 
    \frac{1}{\sigma^2} \mathop{\mathbb{E}_{\mathbf{\epsilon} \sim N(0, \sigma^2 \mathbf{I})}} \left[ F(\mathbf{\theta} + \epsilon) \mathbf{\epsilon} \right]
\end{equation}
which in practice is estimated with samples:
\begin{equation}
\label{eq:es-update}
    \nabla_\mathbf{\theta} F(\mathbf{\theta}) \approx 
    \frac{1}{n \sigma^2} \sum_{i=1}^{n} F(\mathbf{\theta} + \mathbf{\epsilon}_i) \mathbf{\epsilon}_i
\end{equation}

Notice that this reduces to sampling Gaussian perturbation vectors $\mathbf{\epsilon}_i \sim N(0, \mathbf{I})$, evaluating the performance of the perturbed policies and aggregating the results over a batch of samples. The communication overhead between workers is drastically reduced by sharing random seeds, resulting in a highly parallelizable method.

\section{IW-ES: Importance Weighted Evolution Strategies}

ES samples large batches of data, in the order of thousands of trajectories, which are discarded after performing a single policy update. When coupled with SGD and small step sizes, this translates into a poor data efficiency. Such inefficiency is found in most on-policy RL methods, which are unable to leverage previous experience once the policy is updated.

Inspired by the multiple SGD updates per batch of experience in PPO~\cite{schulman2017_ppo}, we propose to modify the ES algorithm to perform several updates to the policy before moving on to collecting a new batch of experience. Should each of these updates be small, it is likely that the population distributions before and after the update will have some overlap, thus making it possible to take more advantage of previous computations and reducing the number of interactions with the environment.

\subsection{Formulation}

Let $\theta^{t} \in \mathbb{R}^{|\theta|}$ denote the population mean after $t$ updates, and $\epsilon_{i}^{t} \in \mathbb{R}^{|\theta|}$ denote the perturbations for which we computed fitness scores, $F(\theta^{t} + \epsilon_{i}^{t})$. We can reuse those samples to update $\theta^{t+k}$ by relying on Importance Sampling to account for the discrepancy between the distribution of the current population and the distribution from which we are actually sampling:
\begin{equation}
    \nabla_\theta F(\theta) \approx 
    \frac{1}{\sigma^2 \sum_{i} c_{i}} \sum_{i=1}^{n} F(\theta^{t} + \epsilon_{i}^{t}) (\theta^{t} + \epsilon_{i}^{t} - \theta^{t+k}) c_i
\end{equation}
where $c_i \in \mathbb{R}$ is the importance weight for the $i$-th perturbation vector. For perturbations drawn from a multivariate Gaussian distribution with diagonal covariance matrix, the computation of $c_i$ can be decomposed as follows:
\begin{equation}
\label{eq:is_weight}
    c_i = 
    \frac{N(\theta^{t} + \epsilon_{i}^{t} - \theta^{t+k}; 0, \sigma^2 \mathbf{I})}{N(\epsilon_{i}^{t}; 0, \sigma^2 \mathbf{I})} = 
    \frac{\prod_{j=1}^{|\theta|} N(\theta_j^{t} + \epsilon_{i,j}^{t} - \theta_j^{t+k}; 0, \sigma^2 )}{\prod_{j=1}^{|\theta|} N(\epsilon_{i,j}^{t}; 0, \sigma^2)}
\end{equation}

This process can be repeated iteratively for $k=(0, \ldots, K)$, updating the policy up to $K+1$ times before collecting a new batch of experience\footnote{The first update for each batch always reduces to the original gradient estimate in ES (Equation~\ref{eq:es-update}), as $\theta^{t+k}=\theta^{t}$ for $k=0$. This is followed by $K$ importance weighted updates.}. In this manuscript we consider $K$ as a fixed hyperparameter, although future work will study strategies that optimally adapt $K$ for each batch.

\subsection{Scalability analysis}
\label{sec:scalability}
One of the most appealing features of ES is its almost perfect scalability to hundreds of CPU cores, and any modification to the original method should retain such property. We base our analysis on the code released by Salimans et al.~\cite{salimans2017_es}, which uses a master-worker architecture. The master broadcasts the parameters at the beginning of each iteration, and the workers send back returns after running rollouts with perturbed versions of the policy. 

The proposed method requires the computation of importance weights, which has a complexity of $\mathcal{O}(batch\_size \cdot |\theta|)$. If those computations are performed sequentially in the master, the time taken by sequential operations might eclipse the benefits of distributing the rollouts across hundreds of workers. This issue can be alleviated by parallelizing the computation of importance weights across all cores in the node hosting the master process. This was enough to provide a throughput close to the baseline method in most of our experiments, but setups with larger models or batch sizes might benefit from a higher level of parallelization. In that case, the computation of importance weights can be distributed across all workers just like the rollouts are: the master broadcasts the updated parameter vector, and the workers send back the scalar importance weights. Note that this incurs in a very little communication overhead, which is key to achieve an efficient distributed computation.

Another implementation trick that can accelerate the computation of importance weights consists in computing $N(\epsilon_{i,j}^{t}; 0, \sigma^2)$ for all possible perturbations at the start of training, trading off memory for computation. It takes advantage of the fact that each worker instantiates a large block of Gaussian noise at the start of training, and $\epsilon_i$ is obtained by sampling $|\theta|$ consecutive parameters at a random index in the noise block. This trick might provide important savings for large models, as the computation of the denominator in Equation~\ref{eq:is_weight} becomes $\mathcal{O}(1)$ instead of $\mathcal{O}(|\theta|)$.

\section{Experiments}

We implement our method on top of the code released by OpenAI\footnote{\url{https://github.com/openai/evolution-strategies-starter/}}.
All experiments run on 720 CPU cores, distributed across 15 machines with 48 cores each. The master process runs on a single core, but the computation of importance weights is parallelized across the 48 cores in the node hosting the master process to accelerate the execution.

We evaluate the method on the Ant-v2 environment\footnote{Although we provide an extensive analysis of IW-ES only on Ant-v2, we have observed similar behaviors on other complex environments, e.g.~Humanoid-v2.} in OpenAI Gym~\cite{brockman2016_gym}, which uses the Mujoco physics engine~\cite{todorov2012_mujoco}. We use the default hyperparameters provided by Salimans et al.~\cite{salimans2017_es} unless otherwise stated. The policy is parameterized by a neural network with two hidden layers of 64 units each and a linear layer that emits continuous actions. Hidden layers are followed by \textit{tanh} non-linearities. Importance weights are clipped at $1$ for numerical stability~\cite{wawrzynski2009_real, munos2016_retrace}. Following previous works~\cite{salimans2017_es, conti2017_novelty}, we evaluate the median reward over \textasciitilde30 stochastic rollouts at each iteration. All reported results are the average over five different runs.

\subsection{Effect of the number of IW updates}
\label{sec:is_updates}

\begin{figure}[ht]
    \centering
    \begin{subfigure}[b]{0.49\textwidth}
        \centering\includegraphics[width=\textwidth]{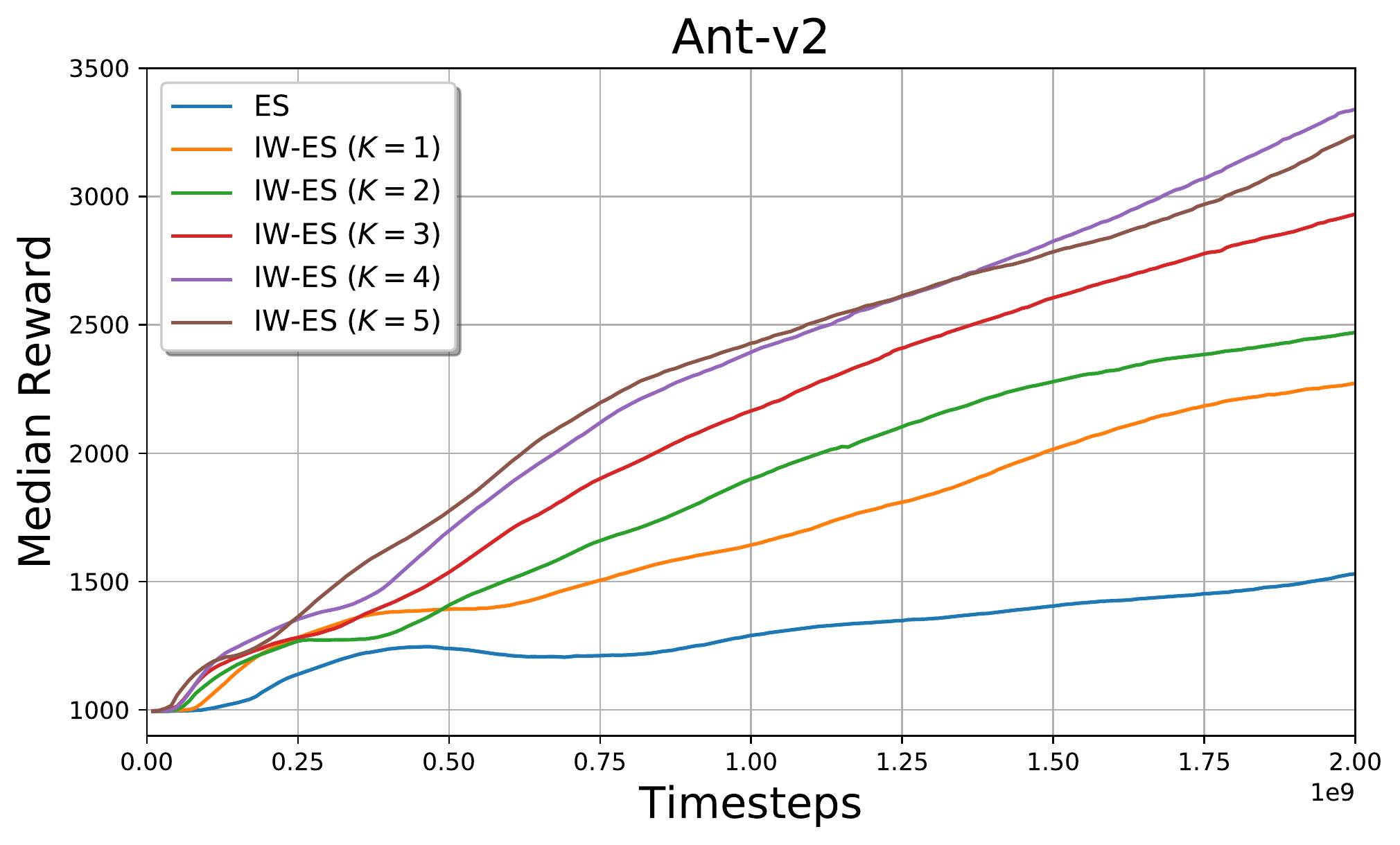}
        \caption{}
        \label{fig:lr4_timesteps}
    \end{subfigure}
    \begin{subfigure}[b]{0.49\textwidth}
        \centering\includegraphics[width=\textwidth]{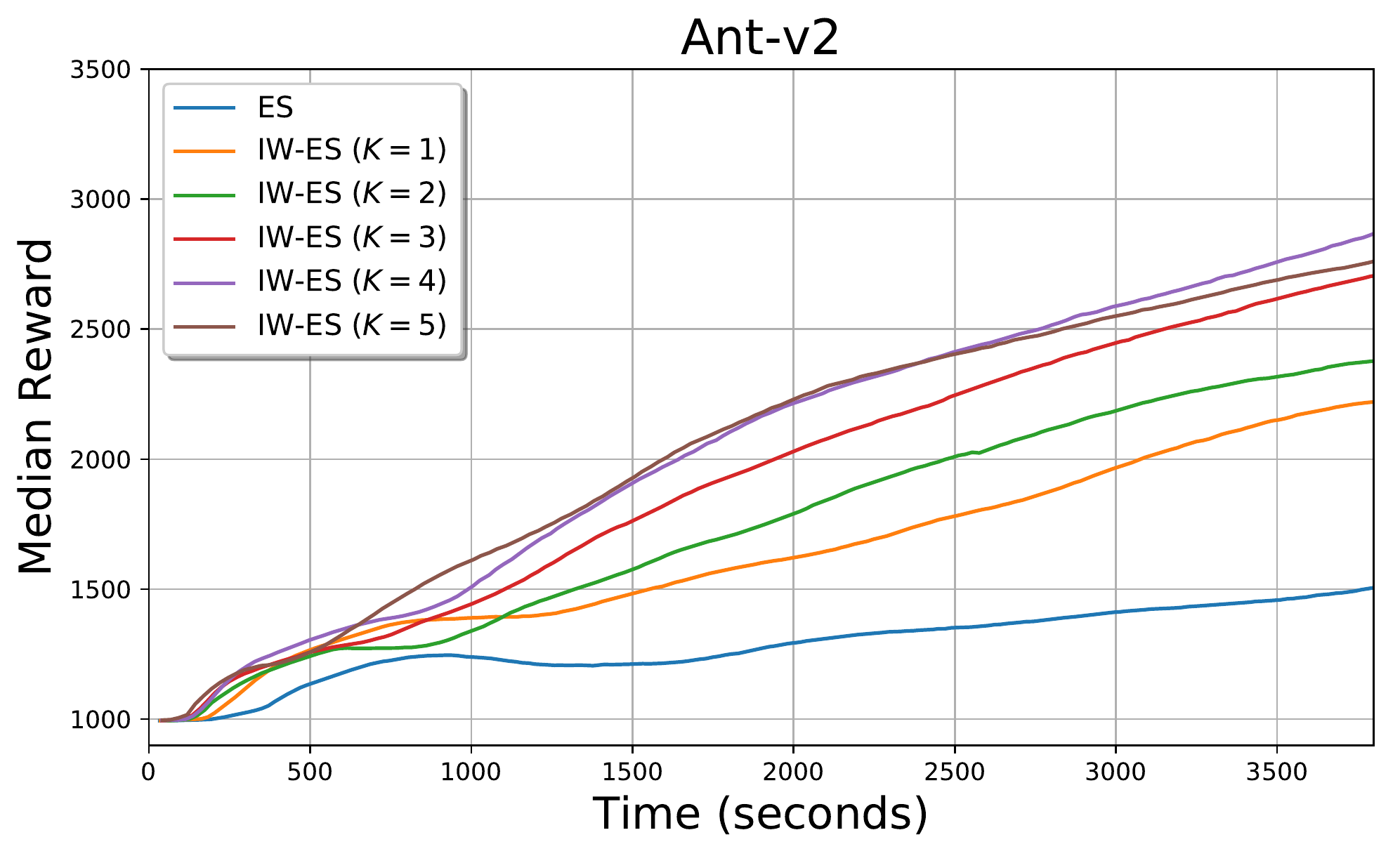} 
        \caption{}
        \label{fig:lr4_seconds}
    \end{subfigure}
    \caption{Performance of ES and IW-ES as a function of \textbf{(a)} the number of interactions with the environment, and \textbf{(b)} wall-clock time. $K$ denotes the number of additional importance weighted updates after each standard update. We observe that additional updates increase the data efficiency of the method in the low learning rate regime, but performing too many importance weighted updates can be detrimental due to an increased variance, e.g.~$K=5$ underperforms $K=4$. A similar trend is observed in terms of wall-clock time.}
    \label{fig:res-lr4}
\end{figure}

The proposed method relies on a high overlap between the population distributions before and after each update, otherwise the variance of the Importance Sampling estimate might become excessively large. For this reason, we first evaluate the effectiveness of additional updates using a low learning rate of $10^{-4}$ that prevents large updates to the policy parameters. As depicted in Figure~\ref{fig:lr4_timesteps}, we observe that additional importance weighted updates provide a faster convergence for a given budget of interactions with the environment. Increased data efficiency also translate in shorter wall-clock times thanks to a reduced computational overhead (Figure~\ref{fig:lr4_timesteps}). However, performance does not always improve when increasing $K$, e.g.~setting $K=5$ instead of $K=4$ results in a performance degradation. This behavior is likely caused by an increased variance in the importance weighted updates for large values of $K$. These results suggest that IW-ES might benefit from strategies that adapt $K$ for each iteration, omitting updates with excessive variance.

\subsection{Effect of the model size}

\begin{figure}[ht]
    \centering
    \begin{subfigure}[b]{0.49\textwidth}
        \centering\includegraphics[width=\textwidth]{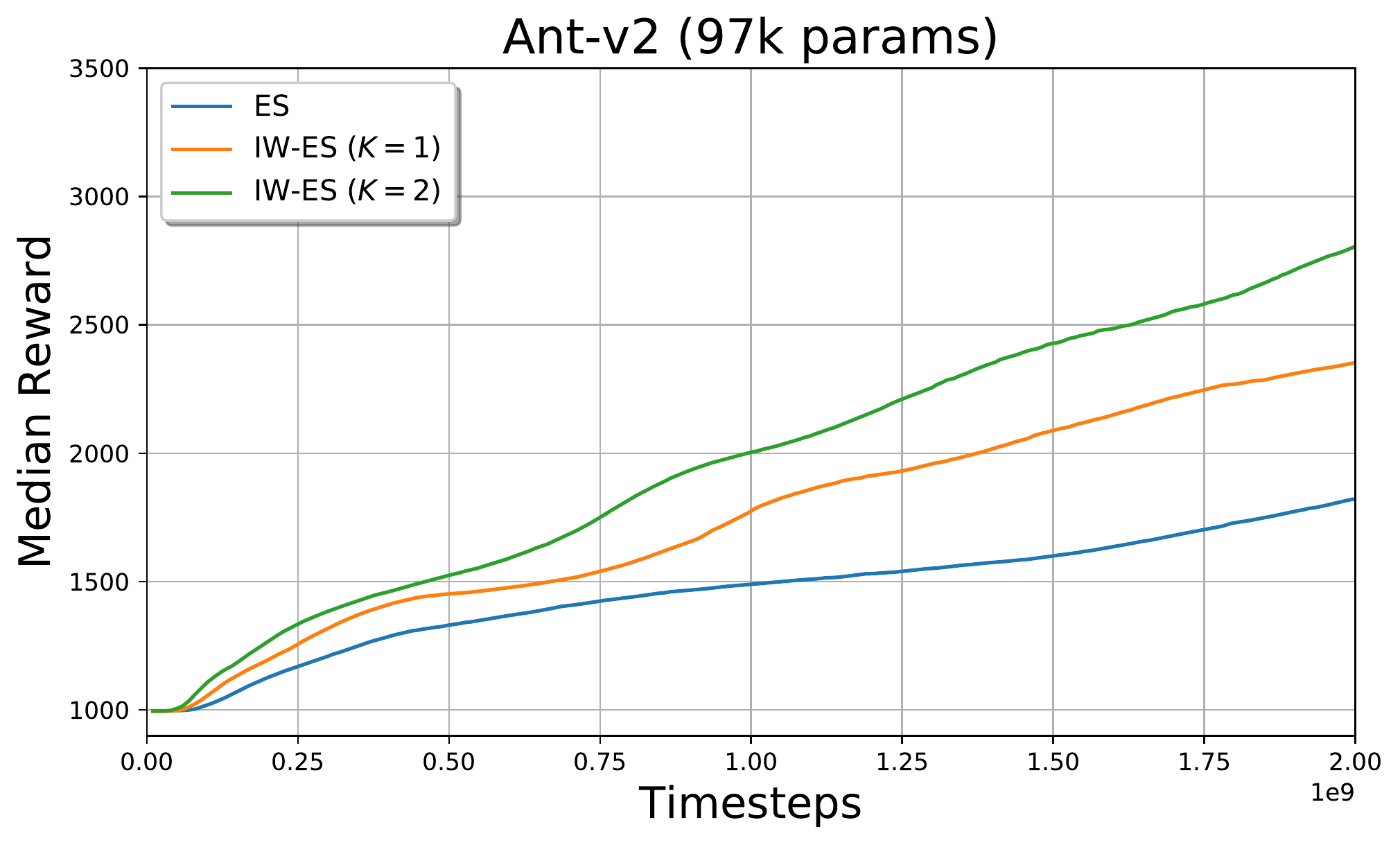}
        \caption{}
        \label{fig:lr4_medium}
    \end{subfigure}
    \begin{subfigure}[b]{0.49\textwidth}
        \centering\includegraphics[width=\textwidth]{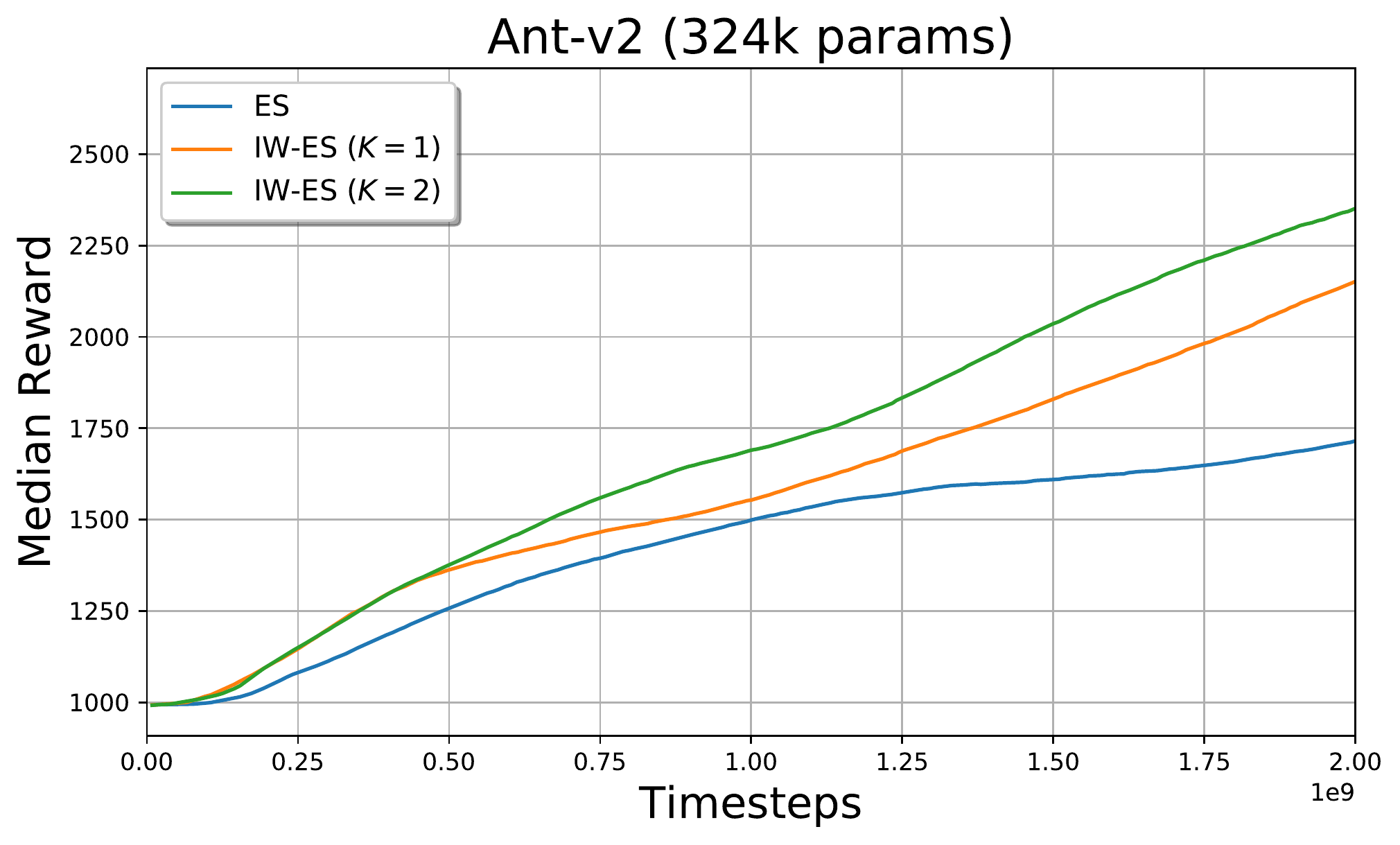}
        \caption{}
        \label{fig:lr4_large}
    \end{subfigure}
    \caption{Performance of ES and IW-ES for larger networks with \textbf{(a)} 256 units per hidden layer, and \textbf{(b)} 512 units per hidden layer.}
    \label{fig:res-model-size}
\end{figure}

A potential source of instability for the proposed method is the computations of importance weights for large models, as they might approach zero or infinity much faster for large values of $|\theta|$ (see Equation~\ref{eq:is_weight}). We experimentally evaluate whether this hinders the performance of IW-ES by training larger networks, with 256 and 512 units in each hidden layer. These larger models have 97k and 324k parameters, respectively, whereas previous experiments considered a much smaller network with 12k parameters. Results reported in Figure~\ref{fig:res-model-size} suggest that IW-ES is robust to the number of parameters in the model, as the benefit of adding additional updates per batch are similar to those observed for smaller models.

\begin{figure}[htb]
    \centering\includegraphics[width=0.7\textwidth]{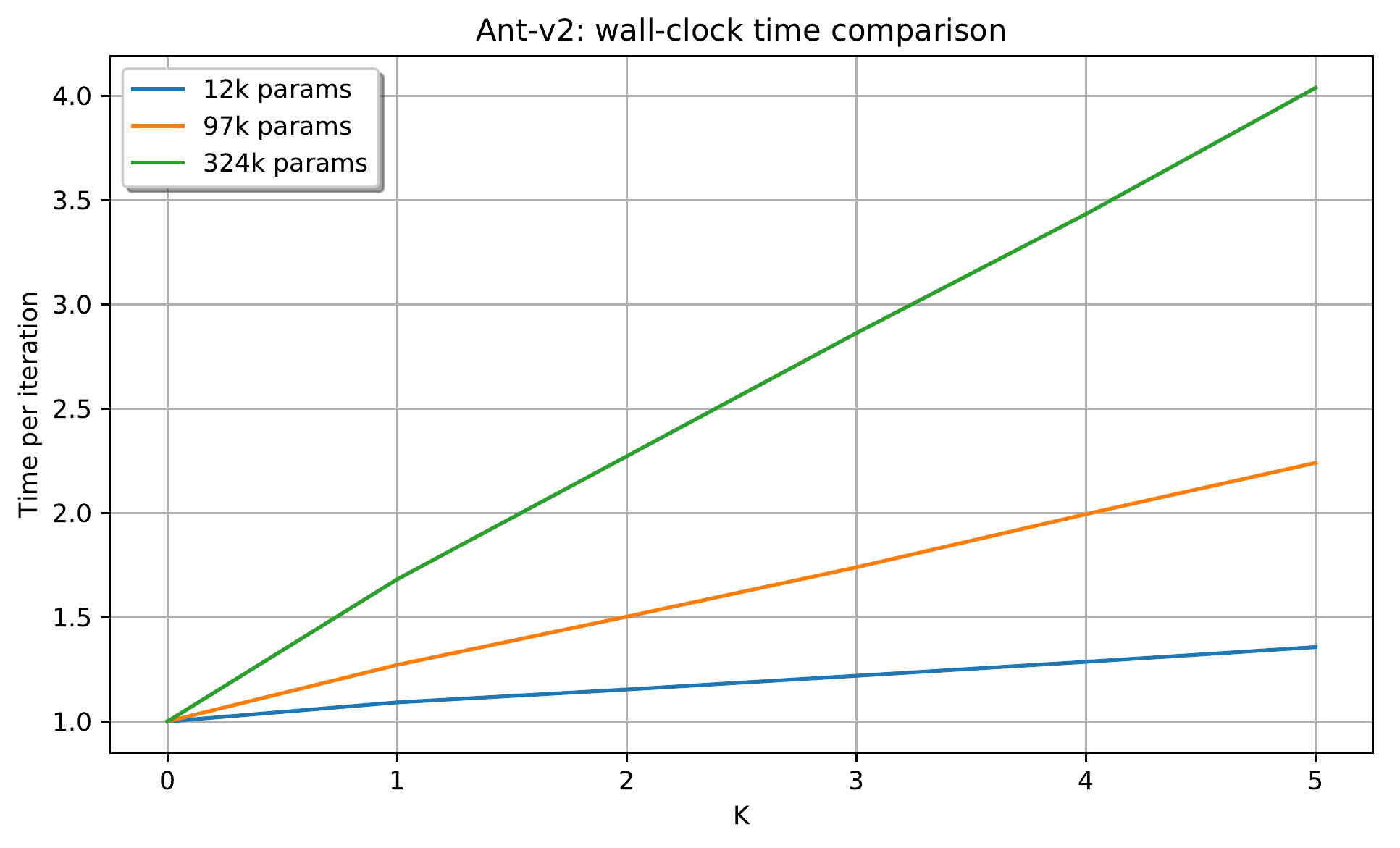}
    \caption{Time per iteration for different values of $K$, normalized by the time taken by ES (i.e.~$K=0$). Our implementation parallelizes the computation of importance weights only across the CPU cores in the node hosting the master process, which becomes a bottleneck for larger models.}
    \label{fig:scalability}
\end{figure}

Figure~\ref{fig:scalability} shows the throughput degradation introduced by IW-ES for each model size and number of importance weighted updates. Since our implementation only leverages 48 of the 720 available CPU cores for computing the importance weights, such computation becomes a bottleneck for larger models and hinders the scalability of the method. This observation motivates the distributed implementation described in Section~\ref{sec:scalability}, which should accelerate IW-ES considerably for large models thanks to the reduced communication overhead between machines.

\subsection{Effect of the learning rate}

\begin{figure}[htb]
    \centering
    \begin{subfigure}[b]{0.49\textwidth}
        \centering\includegraphics[width=\textwidth]{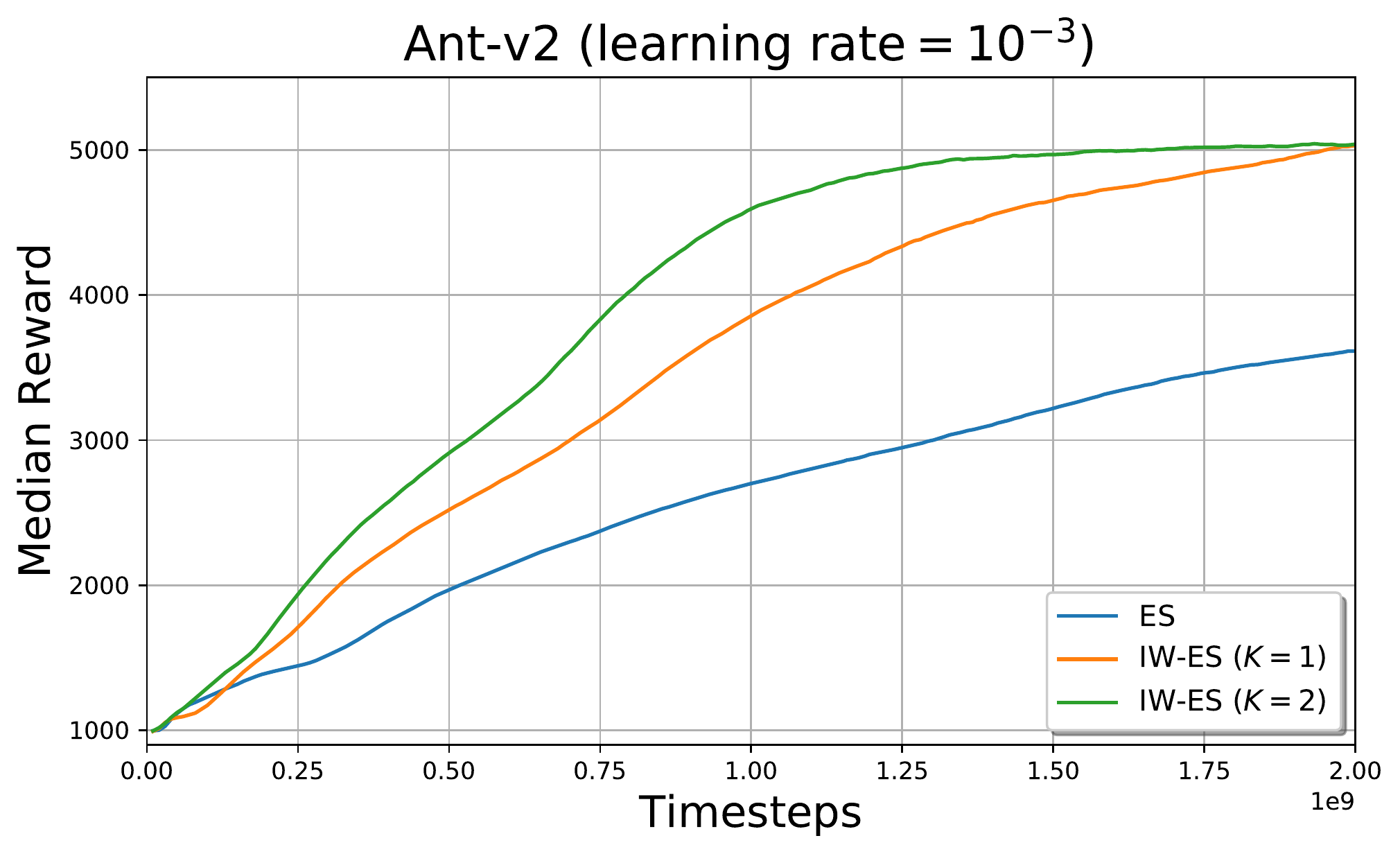}
        \caption{}
        \label{fig:lr3_timesteps}
    \end{subfigure}
    \begin{subfigure}[b]{0.49\textwidth}
        \centering\includegraphics[width=\textwidth]{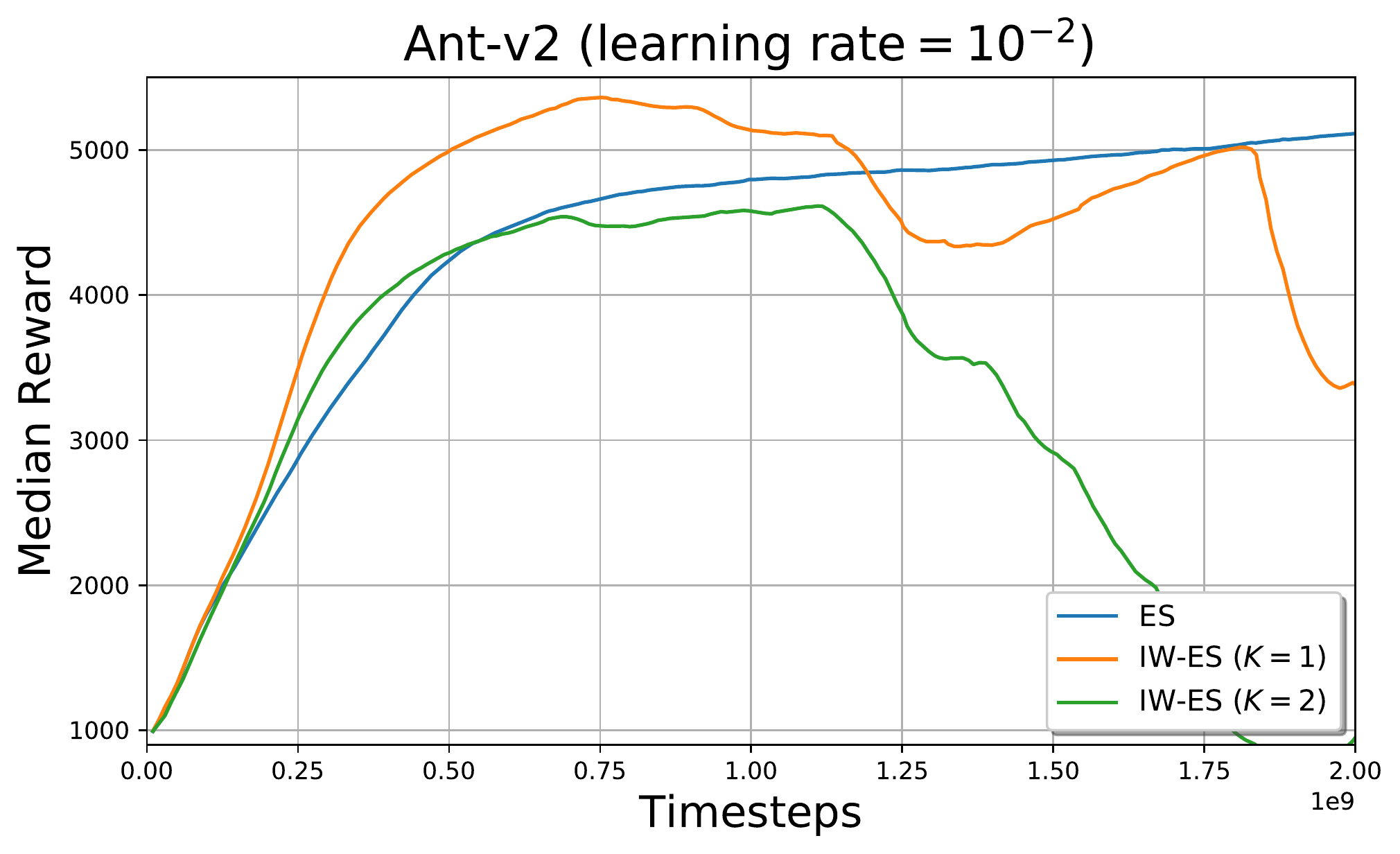} 
        \caption{}
        \label{fig:lr2_timesteps}
    \end{subfigure}
    \caption{Performance of ES and IW-ES with learning rates of \textbf{(a)} $10^{-3}$, and \textbf{(b)} $10^{-2}$. Larger learning rates reduce the benefits of IW-ES due to a larger variance of the Importance Sampling estimate.}
    \label{fig:res-learning-rate}
\end{figure}

ES benefits from larger learning rates than those employed in previous experiments, as they provide faster convergence and thus increased data efficiency, but larger step sizes might increase the variance of IW-ES updates as well due to a larger mismatch between distributions. We evaluate this hypothesis by training policies with larger learning rates of $10^{-3}$ and $10^{-2}$. Results reported in Figure~\ref{fig:res-learning-rate} confirm that importance weighted updates not only become less effective with larger learning rates, but can even become unstable and underperform the baseline ES. 

These experiments consider the learning rate as a proxy for controlling the overlap between the distributions before and after each update, which is the actual measure determining the variance of importance weighted updates. Even though a finer grained search over learning rate values could be carried out in order to determine whether IW-ES can outperform ES under optimal hyperparameters, we argue that next steps should aim at controlling the similarity between the distributions before and after each update. For instance, drawing a parallelism with trust region-based methods~\cite{schulman2015_trpo,heess2017_dppo}, a constraint could be added on the KL divergence between distributions.

\section{Related Work}

Some works have proposed extensions or modifications to the original Evolution Strategies (ES) algorithm proposed by Salimans et al.~\cite{salimans2017_es}. These include an update rule inspired by genetic algorithms~\cite{such2017_deepga} and training a meta-population of agents that optimize both for reward and novelty~\cite{conti2017_novelty}. The possibility of optimizing non-differentiable functions with ES has also allowed to learn loss functions for RL in a meta-learning setup~\cite{houthooft2018_epg}.

The design of data-efficient methods for RL has garnered much research attention, mostly through off-policy methods that can leverage experience collected by policies other than the one being optimized. This advantage, often associated to value-based methods such as Q-learning~\cite{watkins1989_learning, mnih2015_dqn}, usually results in an increased data efficiency. Policy-based methods may also leverage off-policy data by accounting for the discrepancy between the behavior and target policies~\cite{precup2001_eligibility, munos2016_retrace}. PPO~\cite{schulman2017_ppo} performs several SGD updates for every batch of collected experience, using Importance Sampling to leverage data collected by an outdated version of the policy, in a similar fashion to our IW-ES update rule.

The IMPALA architecture~\cite{espeholt2018_impala} can scale training of actor-critic methods across many machines to achieve a high troughput, enabling advances in multi-task RL~\cite{hessel2018_multi}. This is achieved by a combination of algorithmic and engineering advances. Its main drawback comes from a fairly uncommon hardware setup, with each GPU paired with over 100 CPU cores, that may not be feasible to put together within many organizations. In comparison, ES requires from less engineering efforts to achieve high troughputs, thanks to the reduced communication overhead, and its hardware requirements are generally easier to meet.

\section{Conclusion \& Future Work}



We introduced Importance Weighted Evolution Strategies (IW-ES), a variant of Evolution Strategies (ES)~\cite{salimans2017_es} that can perform several model updates with a single of batch of data. Under the desired conditions, i.e.~when samples from the population distribution before the update are still likely under the updated distribution, IW-ES demonstrated a higher data efficiency than that of ES. For small models, these benefits can be introduced with a small increase in sequential computational load that maintains the scalability of ES. For larger models, we describe how to leverage distributed hardware to distribute further parallelize the added computation and achieve higher throughput rates.

Besides implementing the completely distributed version of IW-ES that can make the most of the available hardware, future work will focus on making IW-ES more resilient to large divergences between distributions that increase the variance of the Importance Sampling estimates. First, an adaptive strategy for $K$ can be designed so that importance weighted updates are made only when their variance is sufficiently low. On the other hand, controlling the divergence after an update through a constraint in the training objective can make IW-ES more robust for large learning rates, and avoid the collapse observed in some experiments. Although applied to policy space instead of parameter space, similar motivations have led to more efficient and stable policy gradient methods~\cite{schulman2015_trpo, schulman2017_ppo}. These lines of research may also lead to revisiting the role of $\sigma$, which controls the spread of the perturbation vectors in ES, but also plays an important role in determining the importance weights in IW-ES.

\section*{Acknowledgments}


We would like to thank Francesc Sastre for his help in deploying the code in MareNostrum, as well as the technical support team at the Barcelona Supercomputing Center. 

This work was partially supported by the Spanish Ministry of Economy and Competitivity and the European Regional Development Fund (ERDF) under contracts TEC2016-75976-R and TIN2015-65316-P, by the BSC-CNS Severo Ochoa program SEV-2015-0493, and grant 2014-SGR-1051 by the Catalan Government.
V{\'\i}ctor Campos was supported by Obra Social ``la Caixa'' through La Caixa-Severo Ochoa International Doctoral Fellowship program.

\bibliographystyle{plain}
\bibliography{references.bib}

\end{document}